\documentclass[11pt, a4paper]{elsarticle}
\usepackage[utf8]{inputenc}
\usepackage[margin=1.5cm]{geometry}
\usepackage{titlesec}
\usepackage{tabu}
\usepackage{longtable}
\usepackage{enumitem}
\usepackage{amssymb}
\usepackage{xcolor}
\newlist{selectlist}{itemize}{2}
\setlist[selectlist]{label=$\square$,leftmargin=*,noitemsep,topsep=0pt}

\usepackage{lmodern}
\usepackage{hyperref}
\hypersetup{
    colorlinks=false,
    linkcolor=blue,
    filecolor=magenta,      
    urlcolor=blue,
}
\usepackage[binary-units=true]{siunitx}
\usepackage{dirtree}
\usepackage{multicol}

\titleformat{\section}[block]{\hspace{1em}\bfseries}{\thesection.}{0.5em}{} 
\titleformat{\subsection}[block]{\hspace{1em}}{\thesubsection}{0.5em}{}

\begin{document}
\begin{flushleft}

\textbf{Article information}\\
\vskip 0.5cm
\textbf{Article title}\\ 
A Dataset of Images of Public Streetlights with Operational Monitoring using Computer Vision Techniques
\vskip 0.5cm
\textbf{Authors}\\ 
Ioannis Mavromatis, Aleksandar Stanoev, Pietro Carnelli, Yichao Jin, Mahesh Sooriyabandara, Aftab Khan
\vskip 0.5cm
\textbf{Affiliations}\\ 
Bristol Research and Innovation Laboratory, Toshiba Europe Ltd., 32 Queen Square, Bristol, BS1 4ND, United Kingdom
\vskip 0.5cm
\textbf{Corresponding author’s email address and Twitter handle}\\ 
\{Ioannis.Mavromatis, Aleksandar.Stanoev, Pietro.Carnelli, Yichao.Jin, Mahesh, Aftab.Khan\}@toshiba-bril.com
\vskip 0.5cm
\textbf{Keywords}\\ 
Streetlight, Street Furniture, Light and Emergency Light Maintenance, Lighting Columns, Computer Vision, Machine Learning, Smart Cities
\newpage\textbf{Abstract}\\ 
A dataset of street light images is presented. Our dataset consists of $\sim350\textrm{k}$ images, taken from 140 UMBRELLA nodes installed in the South Gloucestershire region in the UK. Each UMBRELLA node is installed on the pole of a lamppost and is equipped with a Raspberry Pi Camera Module v1 facing upwards towards the sky and lamppost light bulb. Each node collects an image at hourly intervals for \SI{24}{\hour} every day. The data collection spans for a period of six months.

\vskip 0.2cm

Each image taken is logged as a single entry in the dataset along with the Global Positioning System (GPS) coordinates of the lamppost. All entries in the dataset have been post-processed and labelled based on the operation of the lamppost, i.e., whether the lamppost is switched ON or OFF. The dataset can be used to train deep neural networks and generate pre-trained models providing feature representations for smart city CCTV applications, smart weather detection algorithms, or street infrastructure monitoring. The dataset can be found at \url{https://doi.org/10.5281/zenodo.6046758}.

\vskip 0.5cm

\textbf{Keywords:} Streetlight, Street Furniture, Light and Emergency Light Maintenance, Lighting Columns, Computer Vision, Machine Learning, Smart Cities

\vskip 0.5cm
\newpage\textbf{Specifications table}\\
\vskip 0.2cm 
%
%
\renewcommand*{\arraystretch}{1.4}\begin{longtable}{|p{33mm}|p{124mm}|}
\hline
\textbf{Subject}                & Computer Science\\
\hline                         
\textbf{Specific subject area}  & Computer Vision and Pattern Recognition \\
\hline
\textbf{Type of data}           & The image data for each street light column are provided as JPEG files~\cite{jpeg}. Each zipped directory contains all the all images associated to a single light column/lamppost. A CSV file containing the information about all the street lights, their metadata and the status decision outcome from our post-processing algorithm, i.e., if it is switched ON or OFF. We also include another processed CSV file with the number of occurrences per street light column for easier post-processing of the data.
\\                                   
\hline
\textbf{How the data were acquired} & Acquired from the UMBRELLA testbed~\cite{UMBRELLA}, consisting of 140 UMBRELLA nodes, each equipped with a Raspberry Pi Camera Module v1~\cite{picamera} and an OmniVision OV5647 sensor~\cite{camerasensor}. The nodes are installed on top of lampposts, around the South Gloucestershire region in the UK, across a $\sim$7.2km stretch of road and the University of West of England Campus. Note, all cameras are facing upwards, towards the light bulbs of the lampposts and the sky.
\\
\hline                         
\textbf{Data format}            & Raw data format, as recorded by each UMBRELLA node, pertaining to the UMBRELLA testbed, in the form of JPEG and CSV files. The raw data are analysed to identify the street lights functionality and the results are listed in a CSV file format.
\\                                                    
\hline
\textbf{Description of          
data collection}             & The data has been recorded on hourly intervals at a best effort basis (for all available nodes on each hourly timeslot), for seven days per week and for a period of six months. A random time delay of \SIrange{0}{600}{\second} was introduced before taking each photo to spread out the analysis resource utilisation on the server side.
\\                         
\hline                         
\textbf{Data source location}   & Roughly $\sim80\%$ of the nodes installed on a $\sim$\SI{7.2}{\kilo\meter} stretch of public road in the Gloucestershire region in the UK, comprising of the following roads: Coldharbour Lane, Filton Road, and A4174 Ring Road. $\sim20\%$ of the nodes are installed around the University of the West of England Frenchay Campus, in Bristol, UK.

\\
\hline                         
\hypertarget{target1}
{\textbf{Data accessibility}}   & Data available online at the link below: \newline \url{https://doi.org/10.5281/zenodo.6046758}~\cite{dataset_zenodo}
\\                         
\hline                         
\textbf{Related                 
research\newline
article}                & 





\\
\hline                         
\end{longtable}

\newpage

\textbf{Value of the Data}\\
\begin{itemize}
\itemsep=0pt
\parsep=0pt
\item[$\bullet$] Our dataset consists of $\sim$350k JPEG images of street lights, taken from lighting columns installed on public road infrastructure in the city of Bristol, UK. Each UMBRELLA node installed on a street column provides a unique camera placement, photographic angle, and distance from the street light. Furthermore, several street lights are partially obstructed by vegetation. In addition, several light bulbs are outside of the Field of View (FoV) the Pi Camera module achieves. Finally, the cameras facing the sky are susceptible to weather conditions (e.g., rain, snow, direct sun and moon light, etc.) that can partially or entirely alter the quality of the images taken. The above provides a unique and diverse dataset of images that can be used for training tools and Machine Learning (ML) models for inspection, monitoring and maintenance use-cases. For example, in~\cite{flStreetLightApplication} we see an example of using this data for evaluating the feasibility of using centralised, personalised, and federated learning for a streetcare IoT application.

\item[$\bullet$] By building upon the provided images, experts from the fields of Computer Vision, Internet of Things (IoT), and Smart Cities can develop and train ML models and heuristic tools that assess the status of the street and emergency lights in real-time (e.g., as in~\cite{streetlightmonitoring}). For instance, this can facilitate further research on the provision of Smart City services that can automatically detect whether a street light is ON or OFF, alert a maintenance team, and reduce the person-hours required for on-site monitoring.

\item[$\bullet$] Deep neural network training and the regularisation of the models produced can benefit from pre-training models~\cite{pretraining}. Our dataset provides an extensive collection of images taken at different times of the day and night, different weather conditions and different exposure settings. ML models pre-trained on this dataset can provide excellent feature representations in deep neural network-based transfer learning applications, e.g., outdoor smart city CCTV deployments.

\item[$\bullet$] Since the cameras in question are facing the sky, this dataset can also be used for training real-time weather warning and monitoring systems. For instance, building a simple rain detector is straightforward, as very different images are produced under rainy conditions. In this use case, the direction of rain can be estimated, and potentially by using existing infrastructure (e.g., existing traffic cameras), weather warnings can be generated, pin-pointed very precisely and disseminated to drivers approaching specific streets.

\item[$\bullet$] Finally, as this dataset provides a very diverse set of streetlamp images, it can be combined with datasets of images from other street furniture (e.g., traffic lights, street name signs, traffic signs, etc.) or sensors (e.g. LiDARs). Such a dataset can be later used to train object recognition tools (sensor fusion) for autonomous vehicles and drones systems and accelerate the research towards autonomous navigation on public road infrastructures, e.g., as in~\cite{autonomousdriving}.

\end{itemize}
\vskip 0.5cm

\textbf{Objective}\\
The objective for generating this dataset was to provide a unique and diverse dataset of images that can be used for inspection, monitoring and maintenance within IoT ecosystems and object recognition use-cases (if combined with other datasets). We designed the dataset in such a way so the images are diverse and unique. This comes from the fact that the camera's FoV, obstructions, and weather conditions affect the image quality providing unique edge use-cases for testing and experimentation. Hence, such a dataset can introduce various challenges when designing detection and classification algorithms, training tools and machine learning models and can benefit researchers in the areas of Computer Vision, Smart Cities, and Machine Learning.

\vskip 0.5cm

\textbf{Data Description}\\
The raw data files associated with each lamppost (namely the JPEG~\cite{jpeg} images and the CSV file), are organised as follows. The dataset is upload as a single Zip file, that, when unzipped, it unfolds to a number of zipped and two CSV files. As our dataset is fairly large, we provide two Zip files, one containing the complete dataset and another one containing an example dataset with a smaller number of JPEG images and UMBRELLA nodes. The naming convention for both is ``streetcare-dataset-complete.zip'' and ``streetcare-dataset-example.zip`'' respectively. The CSV files within the root directory follow a similar naming convention, i.e., ``occurrences-complete.csv'', ``streetlights-complete.csv'' for the complete dataset and ``occurrences-example.csv'', ``streetlights-example.csv'' for the example one.

\vskip 0.2cm

Within the root zipped directory, one can find the Zip files containing our JPEG images. These zipped files are named after the ``serial ID'' of each UMBRELLA node, i.e., a friendly name given to each node. This serial ID is in the form of ``\textit{RS[ES]-[-A-Z0-9]}''. For example, valid serials are ``RSE-A-11-C'', ``RSE-A-G-8-C'', ``RSS-47-C'', etc. All serial IDs represent the type of the node and the sensors equipped, but this is outside of the scope of this dataset. With what regards this dataset, all nodes are equipped with the same Raspberry Pi Camera Module (i.e., Camera Module V1~\cite{picamera}). All the ``RSE-*'' or ``RSS-*'' subdirectories, when unzipped, contain the JPEG images taken of this particular UMBRELLA node. The file name of each JPEG file is a random string of characters and numbers with a size of 32 digits. This string is assigned at random when the file is created and is always unique. An example of the directory tree of our dataset can be seen below:


\begin{multicols}{2}
\dirtree{%
.1 streetcare-dataset-complete.zip.
.2 occurrences-complete.csv.
.2 streetlights-complete.csv.
.2 RSE-6-C.zip.
.3 9397c82e6dafd06435e13552e2e7d2ca.jpg.
.3 9196c26e24a1ed938d8a170939833c79.jpg.
.3 ....
.2 RSE-A-12-C.zip.
.3 d69bb15b397f33b2f706f89926f904a9.jpg.
.3 f1b60e40d3d295a9eee2971ce5e426bc.jpg.
.3 ....
.2 ....
}

\dirtree{%
.1 streetcare-dataset-example.zip.
.2 occurrences-example.csv.
.2 streetlights-example.csv.
.2 RSE-L-1-C.zip.
.3 cf0f229123a6b1ac659a9f51366c469f.jpg.
.3 b8fb64ece07c34a21791afcff7fad72f.jpg.
.3 ....
.2 RSS-12-C.zip.
.3 e27bd4046e8616906f5daf80ed8b628b.jpg.
.3 8aedbdbdf8aab033bc808be3feaeea34.jpg.
.3 ....
.2 ....
}
\end{multicols}

\vskip 0.5cm

With what regards to the CSV files, the ``streetlights-XXX.csv'' contains in a tabular format all the information about the street lights, their metadata and the status decision outcome from our post-processing algorithm, i.e., ON or OFF. The first line of the file shows the label of each column. For the following rows, each row is an ``entry'', i.e., it represents a JPEG image taken from a specific UMBRELLA node at a specific time. The  ``streetlights-XXX.csv'' files are structured in the following format:
\begin{center}
    {\tt Column\_Label\_0;Column\_Label\_1;$\ldots$ Column\_Label\_n}\\
    {\tt Value\_0\_Entry\_0;Value\_1\_Entry\_0;$\ldots$ Value\_n\_Entry\_0}\\
    {\tt Value\_0\_Entry\_1;Value\_1\_Entry\_1;$\ldots$ Value\_n\_Entry\_1}\\
    {...}\\
    {\tt Value\_0\_Entry\_m;Value\_1\_Entry\_m;$\ldots$ Value\_n\_Entry\_m}\\
\end{center}

\begin{table}[t]
\centering
\caption{Definition of the labels of each column of the CSV data files in the dataset.}
    \label{tab:csvlabels}
\vskip 0.1cm

\begin{tabular}{r|l}
\hline
\textbf{Field Name} & \textbf{Definition} \\ \hline \hline
{\tt id} & A unique ID given to each node. It increments by one for each image captured. \\ 
{\tt serial} & The UMBRELLA nodes ``serial ID'' (friendly name). \\ 
{\tt date} & Date and timestamp at the moment an image was captured. \\ 
{\tt hostname} & The UMBRELLA nodes hostname, as seen in the host OS. \\ 
{\tt lat} & Latitude of the position of the current node. \\ 
{\tt lon} & Longitude of the position of the current node. \\ 
{\tt image\_name} & The JPEG image file name (32 digit alphanumber string with ``.jpg'' extension). \\
{\tt fault\_detected} & The post-processed result of the streetlights status (operational or not). \\
{\tt confidence} & The probability the ``fault\_detected'' result is correct. \\
{\tt daynight} & The astrological day or night time cycle. \\
{\tt red} & The pixel intensity as perceived under the ``red'' RGB channel. \\
{\tt green} & The pixel intensity as perceived under the ``green'' RGB channel.  \\
{\tt blue} & The pixel intensity as perceived under the ``blue'' RGB channel.  \\
\hline
\end{tabular}
\end{table}

A detailed explanation of each column found in the ``streetlights-XXX.csv'' file can be found in Tab.~\ref{tab:csvlabels}. With regards to the ``occurrences-XXX.csv'', it contains a list of serial IDs from all nodes in the dataset, and the number of images taken from this specific node. This file is ordered in descending order based on the number of images. As before, the file is in a tabular format with the first line being the labels of the columns, i.e., ``serial'' and ``occurrences'', and later each line is an entry, i.e., a node serial ID and the number of occurrences in the dataset.


\vskip 0.5cm

\textbf{Experimental design, materials and methods}\\

All UMBRELLA nodes are installed across multiple locations in South Gloucestershire region of the UK, these being a $\sim$\SI{7.2}{\kilo\meter} stretch of public road (about $\sim80\%$ of the nodes) and the University of the West of England (UWE) Frenchay Campus (about $\sim20\%$ of the nodes). The nodes locations and the region can be seen in Fig.~\ref{fig:umbrella_network}. All UMBRELLA nodes are connected to our backend servers via a fibre or a wireless (WiFi) interface. Our servers are used to collect all the images taken and store the information in our database.

\vskip 0.2cm

Fig.~\ref{fig:umbrella_node} shows an Umbrella node attached to a lamppost. Each rhombus segment contains custom PCBs, designed and manufactured by Toshiba, accommodating ten sensors (e.g., Bosch BME680, accelerometers, microphones, etc.) and seven network interfaces (Bluetooth, fibre, WiFi, LoRa, etc.). At the top of the node, one can find a Raspberry Pi Camera Module v1~\cite{picamera}, used for taking the street light images. The camera module is equipped with an OmniVision OV5647 sensor~\cite{camerasensor}. All PCBs are connected to a main processing unit (Raspberry Pi 3b+ Compute Module~\cite{RPI_3_COMPUTE}), responsible for controlling the sensors and the camera module, processing the data generated and sending them to the backend servers. 

\vskip 0.2cm

\begin{figure}[t]
    \centering
    \includegraphics[width=1\textwidth]{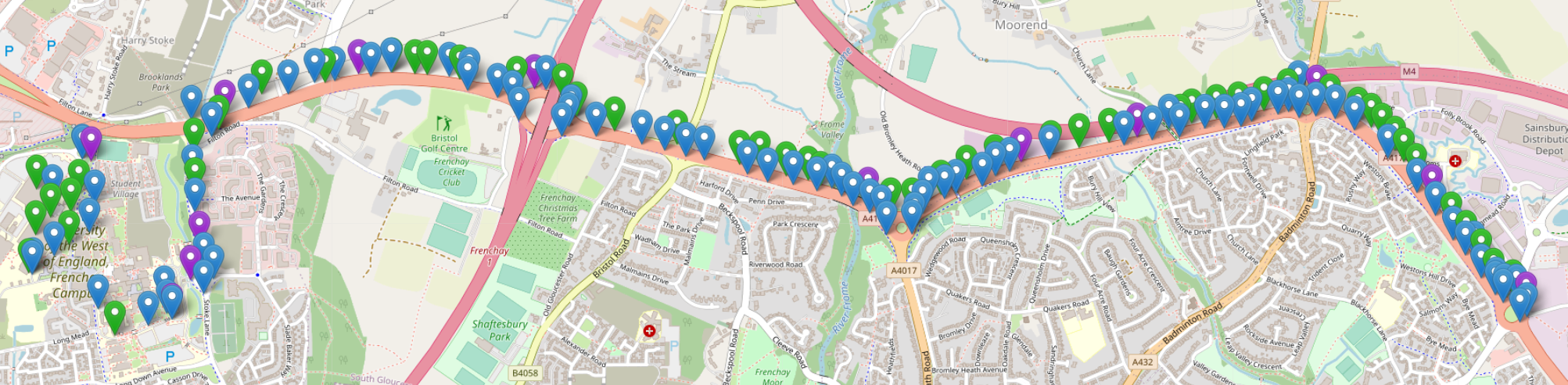}
    \caption{The UMBRELLA network. All nodes are installed on public lampposts across a road of $\sim$\SI{7.2}{\kilo\meter}. The colours represent the nodes connectivity, i.e. green is fibre connected, blue is WiFi connected, and purple is fibre connected and can act as a LoRa gateway too.}
    \label{fig:umbrella_network}
\end{figure}

\begin{figure}[t]
    \centering
    \includegraphics[width=0.80\textwidth]{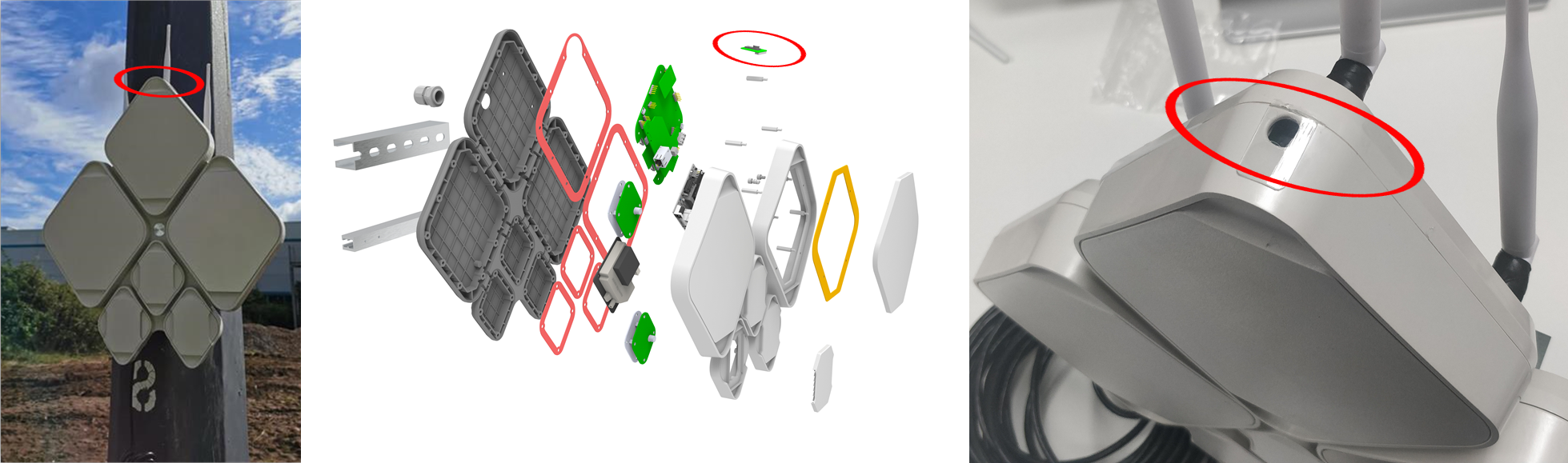}
    \caption{UMBRELLLA Node on a lamppost, its exploded view, and the camera placement at the top of the node (seen in the red circles).}
    \label{fig:umbrella_node}
\end{figure}

All available nodes periodically take a photo of the lamppost and send it back to our backend for processing and storing. The photos are taken at hourly intervals and are collected for the entire \SI{24}{\hour} time period every day. The scheduler works in a best-effort fashion, i.e., if a node is not available on a specific timeslot (e.g., due to downtime), no photos are taken. The photo is scheduled every hour on the hour. A random time delay of \SIrange{0}{600}{\second} is introduced on each node before taking a photo, to ensure the data processing load is spread evenly on the server side. When a photo is saved, an entry is created in our CSV file reporting the ``serial ID'' of the node ({\tt serial} column - Tab.~\ref{tab:csvlabels}), the date and time the photograph was taken ({\tt date} column - Tab.~\ref{tab:csvlabels}), the hostname of the Raspberry Pi the camera was connected to ({\tt hostname} column - Tab.~\ref{tab:csvlabels}), and the GPS latitude and lognitude coordinates of the lamppost ({\tt lat} and {\tt lon} column - Tab.~\ref{tab:csvlabels}). Finally, the JPEG file name is randomly generated using 32 alphanumeric characters and is reported in the CSV file as well ({\tt image\_name} column - Tab.~\ref{tab:csvlabels}).

\vskip 0.2cm

All lampposts operate (are turned ON or OFF) based on the astronomical night time, i.e., they are turned ON \SI{15}{\minute} before the astronomical dusk and are turned OFF \SI{15}{\minute} after the astronomical dawn. Based on that day/night cycle, we change the configuration of our camera accordingly. The different configurations applied to the camera sensor are listed in Tab.~\ref{tab:camerasettings}. During the ``day'' time, i.e., the time that a lamppost is expected to be OFF, a camera is left in automatic exposure and shutter speed settings to compensate for the available sunlight under different weather conditions. During the ``night'' time, i.e., the time that a lamppost is expected to be ON, the exposure, ISO, and shutter speed are increased to compensate for the low-light conditions. During the night time particularly, two images are collected with different configurations. The first one (column ``Night'' in Tab.~\ref{tab:camerasettings}) is the photo that is saved as part our dataset. The second configuration (column ``Night - High exposure'' in Tab.~\ref{tab:camerasettings}) is used for our post-processing and the labelling of the dataset. Two examples of the same photo taken with and without the increased exposure can be seen in Fig.~\ref{fig:highexposure}. The astrological day and night cycle reported for each individual photo taken is reported in the column {\tt daynight} in our CSV file, and is calculated using Astral Python library~\cite{astral}.

\begin{figure}[t]
    \centering
    \includegraphics[width=0.80\textwidth]{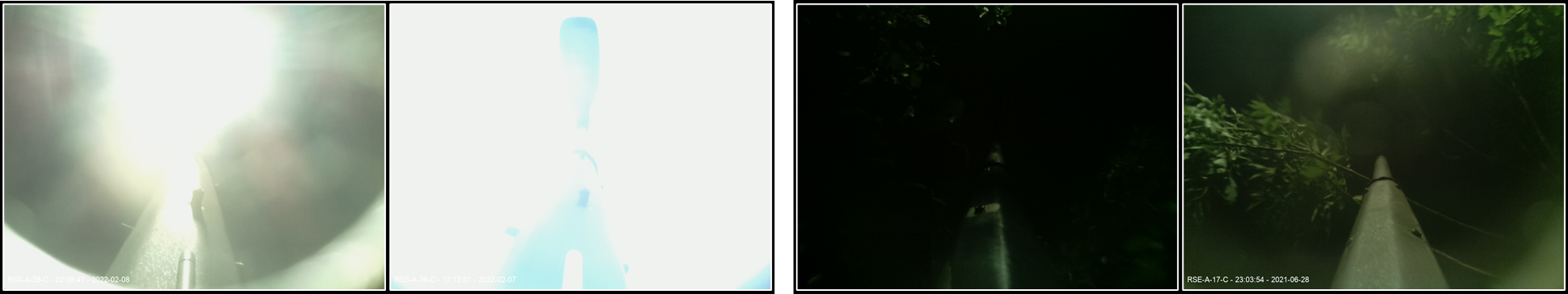}
    \caption{Two examples of the highly exposed photos used for the night time evaluation. On the left, a lamppost with the light directly over the camera. On the right, a lamppost with increased vegetation around the node.}
    \label{fig:highexposure}
\end{figure}

\begin{table}[t]
\centering
\caption{Camera Module and Sensor Settings.}
    \label{tab:camerasettings}
\vskip 0.1cm

\begin{tabular}{r|c|c|c}
\hline
\textbf{Setting} & \textbf{Day} & \textbf{Night} & \textbf{Night - High exposure} \\ \hline \hline
{\tt Exposure Mode} & Automatic & \multicolumn{2}{c}{Manual} \\ \hline
{\tt ISO} & $100$ & $400$ & $800$ \\ \hline
{\tt Shutter Speed} & Automatic & 250000 & 600000 \\ \hline
{\tt LED State} & \multicolumn{3}{c}{Off} \\ \hline
{\tt Flash} & \multicolumn{3}{c}{Off} \\ \hline
{\tt Resolution} & \multicolumn{3}{c}{$1024\times768$} \\ \hline
{\tt Contrast} & \multicolumn{3}{c}{$0$} \\ \hline
{\tt Brightness} & \multicolumn{3}{c}{$50\%$} \\ 
\hline
\end{tabular}
\end{table}

\vskip 0.2cm

Each lamppost image is post-processed on the server-side to identify whether the lamppost is ON or OFF based on its expected behaviour and day/night cycle. More specifically, all photos are converted to a 3-channel Red-Green-Blue (RGB) format with a range of values between \SIrange{0}{255}{}. The values are reported in our CSV file under the {\tt red}, {\tt green} and {\tt blue} columns. During the nighttime, the results reported under the RGB columns come from the highly exposed photos. When the three channels are reported, the streetlight status is calculated, i.e., operational or not ({\tt fault\_detected} column in Tab.~\ref{tab:csvlabels}), as well as a confidence interval ({\tt confidence} column in Tab.~\ref{tab:csvlabels}). As operational is considered a streetlight when it is turned OFF during the day or turned ON during the night cycles. On the other hand, a fault is reported when a streetlight is ON during the day or OFF during the night. An example of how the data are reported and are visualised in our frontend is shown in Fig.~\ref{fig:portal}.

\begin{figure}[t]
    \centering
    \includegraphics[width=0.9\textwidth]{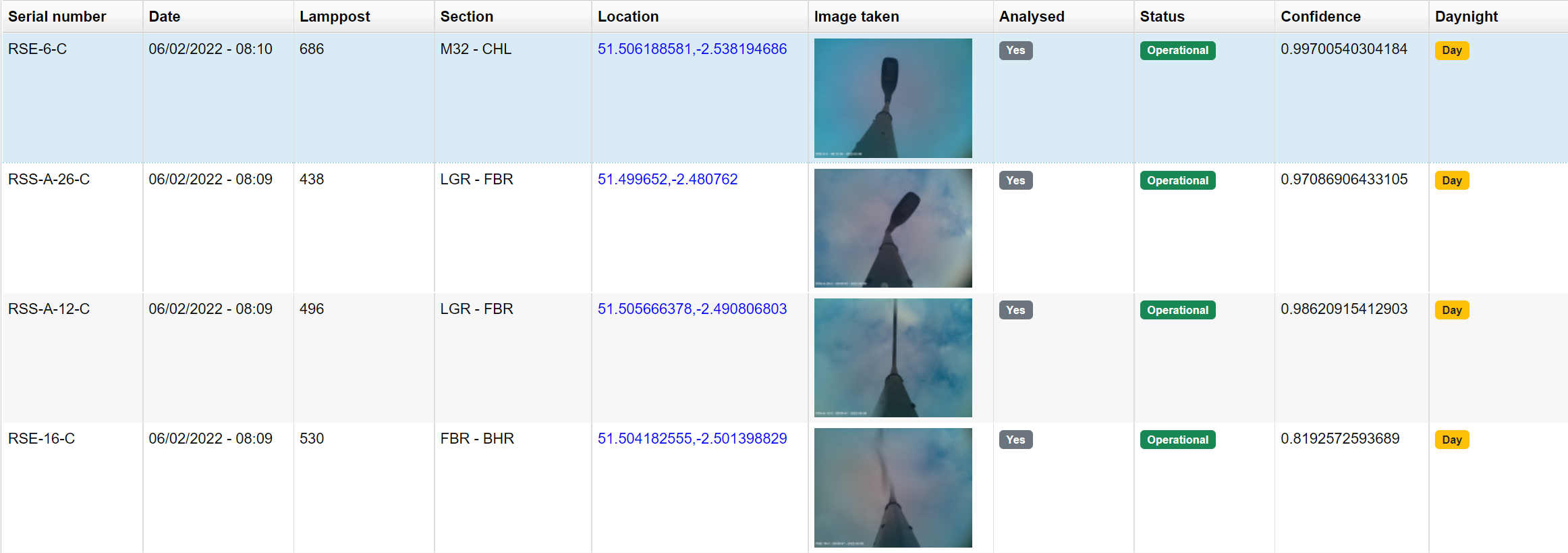}
    \caption{An example of a number of images collected and analysed, as seen on our frontend portal. Their status, the confidence of the prediction, the day/night cycle and information about the nodes (serial ID, location, etc.) are reported along with the photo taken.}
    \label{fig:portal}
\end{figure}

\vskip 0.2cm

With regards the fault detection mechanism, during the day all images are validated against a ML-trained model. Our model is based on a pre-trained VGG-16 model (using the Imagenet dataset). We initially optimised the final layers of the model for the lamppost ON/OFF classification, followed by some fine tuning for the entire model. The output of the model is a Softmax layer working as a binary classifier (ON/OFF). Our ML model operates with an accuracy of circa $\geq90\%$. The neural network architecture can be seen in Fig.~\ref{fig:neural}. More information about the initial VGG-16 model can be found in~\cite{vgg16}. During the night, the RGB values of the node are used for identifying its' status. For the night use-case we use the highly exposed images and the (raw) RGB values. More specifically, when the median RGB value is $\geq200$, the lamppost is considered as ON and the confidence is nearly $1$. On the other hand, when $\textrm{RGB}_{\textrm{median}}\leq100$ the lamppost is considered as OFF and the confidence is $1$ again. For the values $100<\textrm{RGB}_{\textrm{median}}<200$ the green channel is used as a second step for validating the status of the lamppost. It was shown that the images produced by nodes with increased vegetation around them (as in Fig.~\ref{fig:highexposure}-right), are more prevalent on the green channel, thus when the green channel is $\geq200$, then the streetlight is considered as ON. We list these results with a confidence of $0.5$. An example of various lampposts and images labelled during the day or the night time can be seen in Fig.~\ref{fig:allExamples}.

\begin{figure}[t]
    \centering
    \includegraphics[width=0.80\textwidth]{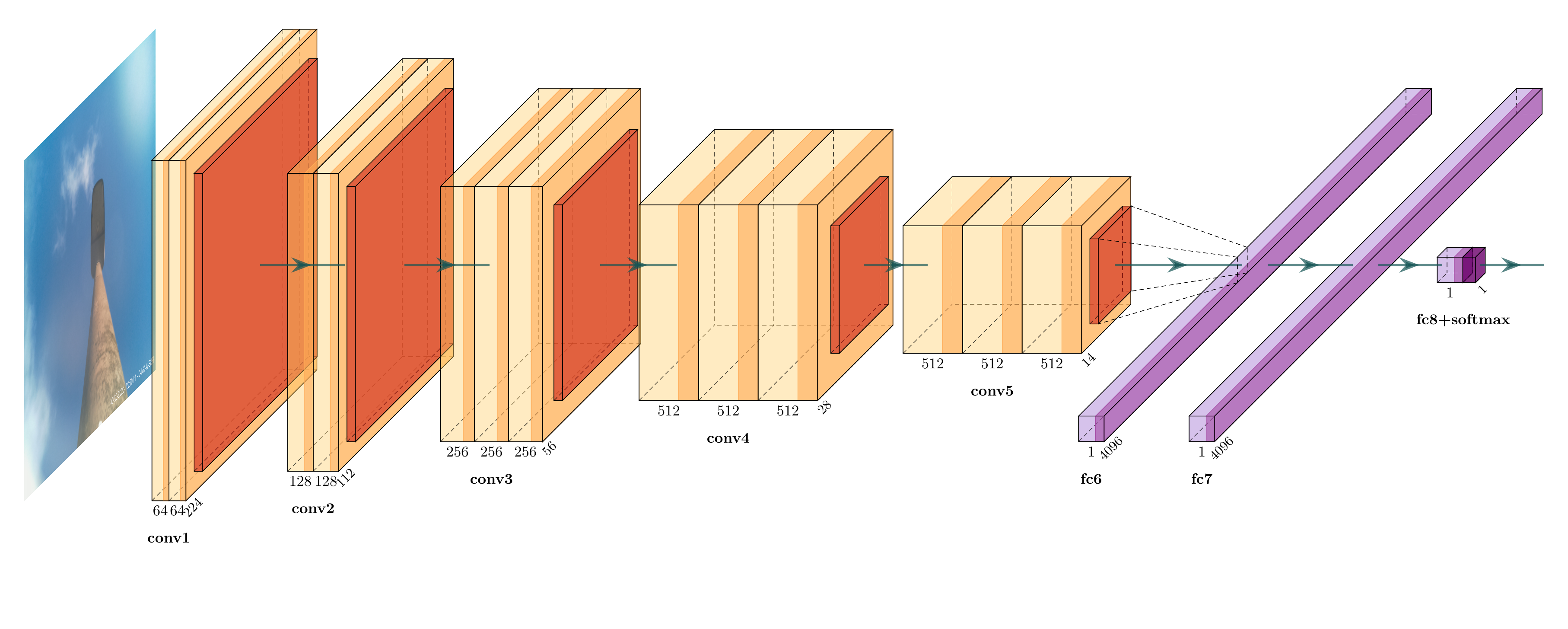}
    \caption{The neural network architecture used for our ``day'' time classification.}
    \label{fig:neural}
\end{figure}

\begin{figure}[t]
    \centering
    \includegraphics[width=0.75\textwidth]{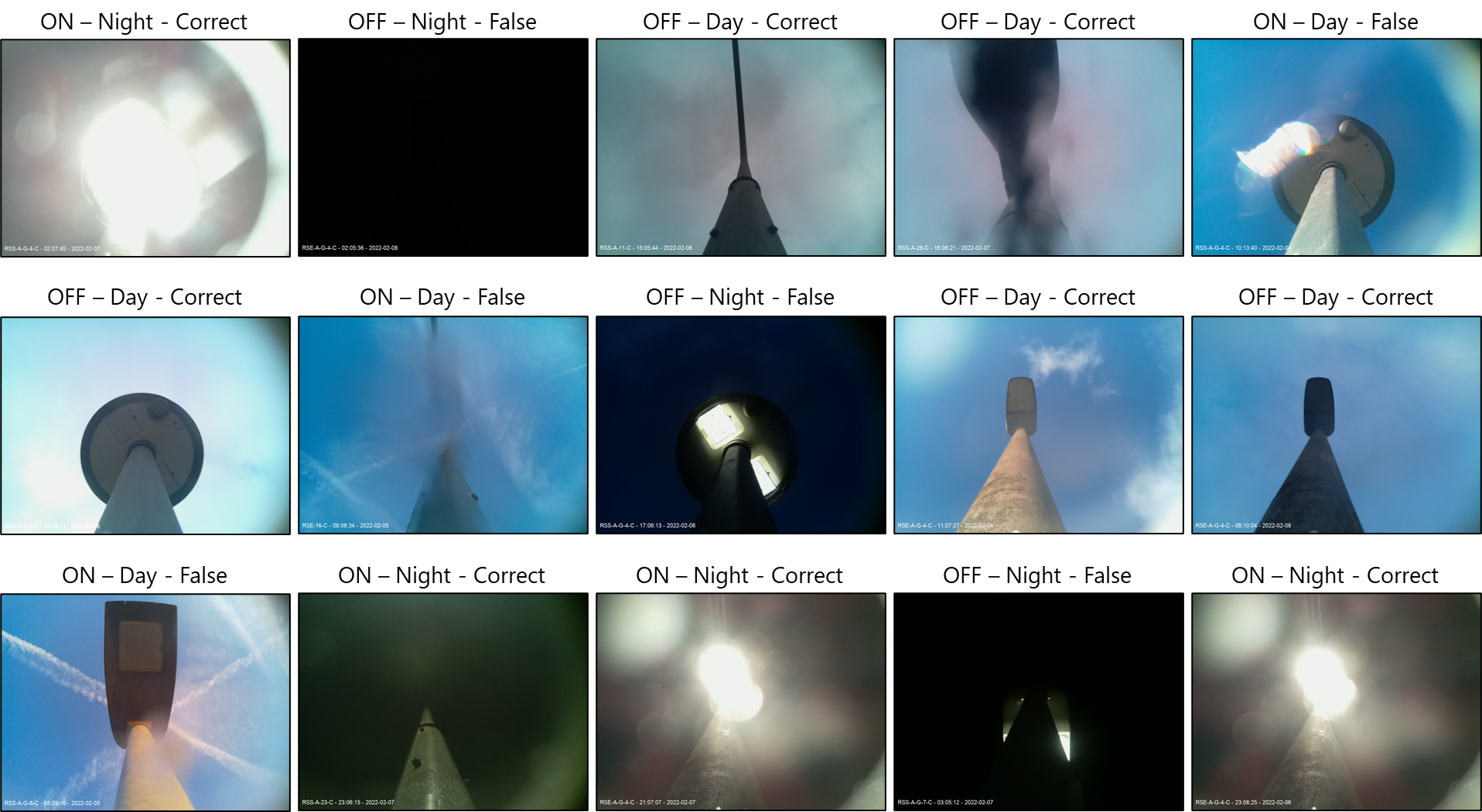}
    \caption{Examples of correctly or falsely labelled photos taken from our system for both ``day'' and ``night'' cases.}
    \label{fig:allExamples}
\end{figure}


\vskip 0.5cm

\textbf{Ethics statements}\\
\noindent 

Hereby, the authors consciously assure that for the manuscript “A Dataset of Images of Public Streetlights with Operational Monitoring using Computer Vision Techniques” the following is fulfilled:
\begin{enumerate}
    \item This material is the authors' own original work, which has not been previously published elsewhere.
    \item The paper is not currently being considered for publication elsewhere.
    \item The paper reflects the authors' own research and analysis in a truthful and complete manner.
    \item The paper properly credits the meaningful contributions of co-authors and co-researchers.
    \item The results are appropriately placed in the context of prior and existing research.
    \item All sources used are properly disclosed (correct citation). Literally copying of text must be indicated as such by using quotation marks and giving proper reference.
    \item All authors have been personally and actively involved in substantial work leading to the paper, and will take public responsibility for its content.
\end{enumerate}

As our dataset did not involve any humal subjects, animal experiments, or social media platform data, approval from an any IRB/local ethics committees was not required. As our camera images are facing the sky, no human subjects are present in the photos. Finally, as our dataset is based on street light images, no survey studies were conducted, and no work was conducted involving chemicals, procedures, or equipment that have any usual hazards inherent in their use, against aminal or human subjects.

I agree with the above statements and declare that this submission follows the policies of Solid State Ionics as outlined in the Guide for Authors and in the Ethical Statement.

\vskip0.5cm

\noindent
\textbf{CRediT author statement}\\
\noindent

\textbf{Ioannis Mavromatis}: Conceptualization, Methodology, Software, Data Curation, Writing - Original Draft, \textbf{Aleksandar Stanoev}: Conceptualization, Methodology, Software, \textbf{Pietro Carnelli}: Software, Writing - Review \& Editing, \textbf{Yichao Jin}: Supervision, \textbf{Mahesh Sooriyabandara}: Funding acquisition, \textbf{Aftab Khan}: Software, Writing - Review \& Editing.

\vskip0.5cm

\textbf{Acknowledgments}\\

This work is funded in part by Toshiba Europe Ltd. UMBRELLA project is funded in conjunction with South Gloucestershire Council by the West of England Local Enterprise Partnership through the Local Growth Fund, administered by the West of England Combined Authority.
\newline

\textbf{Declaration of Competing Interest}\\

\vskip0.3cm
\begin{itemize}
\item[$\checkmark$]{The authors declare that they have no known competing financial interests or personal relationships that could have appeared to influence the work reported in this paper.}

\item[$\square$]{The authors declare the following financial interests/personal relationships which may be considered as potential competing interests: }
\end{itemize}
\vskip0.3cm

\bibliographystyle{IEEEtran}
\bibliography{IEEEabrv,refs}



\end{flushleft}
\end{document}